\documentclass[letterpaper, 10pt, conference]{ieeeconf}  

\IEEEoverridecommandlockouts                              

\usepackage{amsfonts,amssymb} 
\usepackage{subfigure}
\usepackage{booktabs}
\usepackage{diagbox}
\usepackage{graphicx}
\usepackage{multirow}
\usepackage{amsmath}
\usepackage{overpic}
\usepackage{epstopdf}
\usepackage{xcolor}
\usepackage{stfloats}
\usepackage{flushend}

\usepackage{algpseudocode}

\usepackage[linesnumbered,ruled,vlined]{algorithm2e}

\usepackage{caption}
\title{\LARGE \bf 3D Object Detection and Tracking Based on Streaming Data}
\author{Xusen Guo$^{1}$, Jianfeng Gu$^{1}$, Silu Guo$^{1}$, Zixiao Xu$^{1}$,\\ Chengzhang Yang$^{1}$, Shanghua Liu$^{1}$, Long Cheng$^{1,*}$ and Kai Huang$^{1,2}$ 
\thanks{$^{*}$ Corresponding author.}
\thanks{$^{1}$ Xusen Guo, Jianfeng Gu, Silu Guo, Zixiao Xu, Chengzhang Yang, Shanghua Liu, Long Cheng and Kai Huang are with School of Data and Computer Science, the Key Laboratory of Machine Intelligence and Advanced Computing, Ministry of Education, Sun Yat-sen University, Guangzhou, China. (email:\{guoxs3, gujf5, guoslu, xuzx5, yangchzh5, liushh26\}@mail2.sysu.edu.cn, \{chenglong3, huangk36\}@mail.sysu.edu.cn)}
\thanks{$^{2}$ Kai Huang is with Sun Yat-sen University, Shenzhen Institute.}}
\begin{document}
\maketitle

\def\eg{\emph{e.g.}}
\def\Eg{\emph{E.g.}}
\def\etal{\emph{et al. }}
\def\figurename{\emph{Fig. }}
\def\tablename{\emph{TABLE}}

\thispagestyle{empty}
\pagestyle{empty}

\begin{abstract}
Recent approaches for 3D object detection have made tremendous progresses due to the development of deep learning. However, previous researches are mostly based on individual frames, leading to limited exploitation of information between frames. In this paper, we attempt to leverage the temporal information in streaming data and explore 3D streaming based object detection as well as tracking. Toward this goal, we set up a dual-way network for 3D object detection based on keyframes, and then propagate predictions to non-key frames through a motion based interpolation algorithm guided by temporal information. Our framework is not only shown to have significant improvements on object detection compared with frame-by-frame paradigm, but also proven to produce competitive results on KITTI Object Tracking Benchmark, with 76.68\% in MOTA and 81.65\% in MOTP respectively.
\end{abstract}

\section{INTRODUCTION}

3D object detection has received increasing attention in the last few decades due to the growing requirement in robotics and autonomous driving, etc. Compared with 2D image, 3D data can provide accurate locations and shapes information of targets. Current approaches for 3D object detection are mostly carried out in three fronts: image based \cite{7780605, chen20183d}, point cloud based \cite{zhou2018voxelnet,yang2018pixor,simon2018complex}, and multi-view fusion based \cite{chen2017multi,ku2018joint}. Most of these approaches have achieved competitive results but are limited to single frame input.

For most detection tasks, data are obtained in a streaming fashion, thus it is more natural to perform object detection in streaming level. Compared with individual frames, consecutive frames can provide temporal information for long-term prediction of an object, which can reduce noisy detections over time. In addition, the truncated and occluded object can possibly be compensated by the information in its adjacent frames. Therefore, it is important to explore 3D object detection based on streaming data.

Performing 3D object detection on streaming data is however complex. First of all, acquiring accurate 3D information in real world is hard. On one hand, camera data can provide rich appearance features but lack of depth information. On the other hand, though point cloud can precisely provide the position of an object, it is very sparse and thus difficult to represent its appearance. Secondly, how to encode the temporal information in consecutive frames is not obvious. For example, generating 3D scene flow for temporal feature representation will need to determine the correspondence of thousands of points between frames, which is not straightforward and also challenging. Last but not least, due to the sheer numbers of frames in streaming data, frame-by-frame detection will introduce tremendous computational costs, which is impracticable for real-life applications. 

\begin{figure*}
	\begin{center}
		\includegraphics[trim={1.1cm, 3.3cm, 1.5cm, 3.5cm}, clip, width=\textwidth]{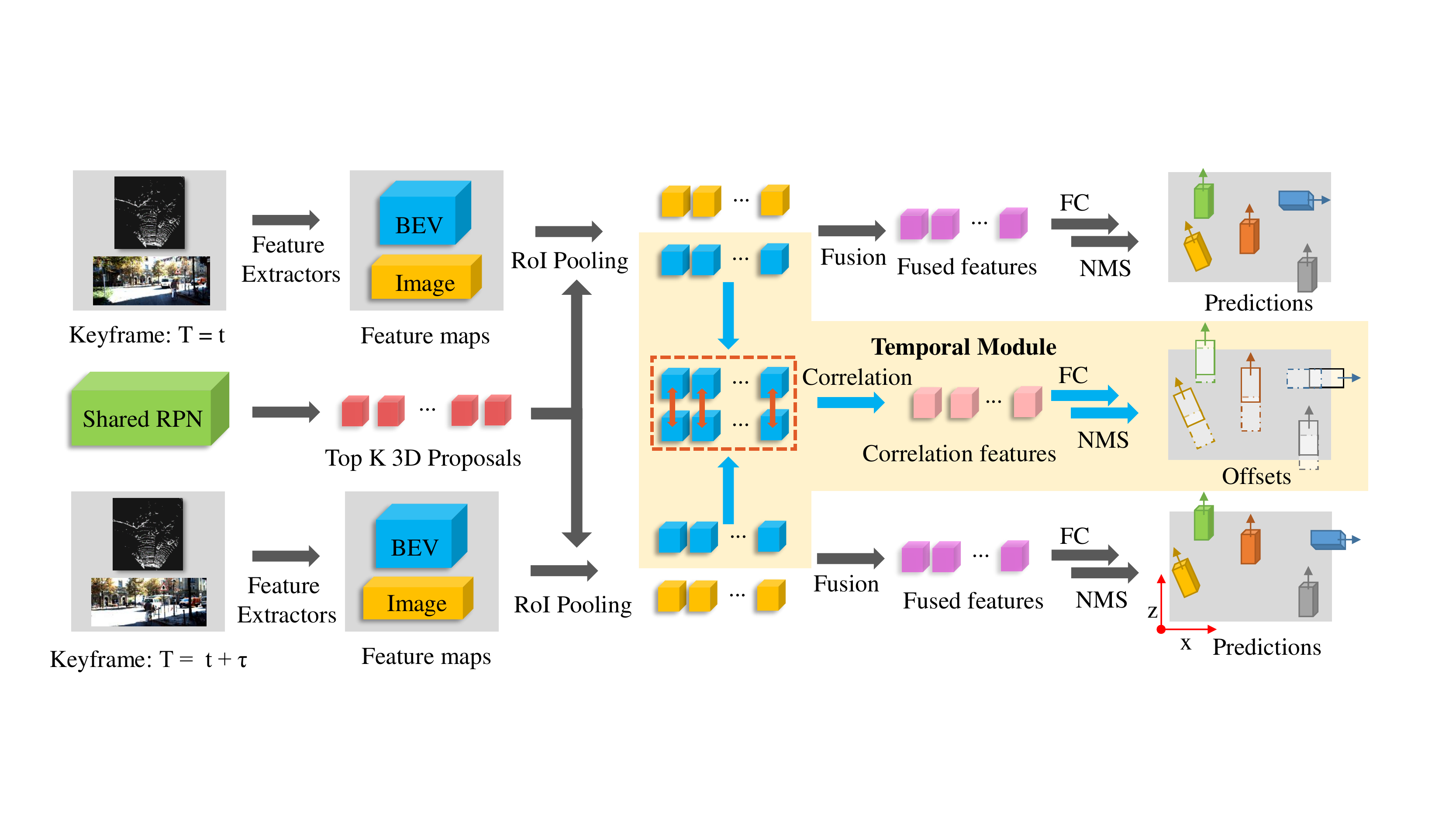}
	\end{center}
	\setlength{\abovecaptionskip}{-1pt}
	\caption{DODT architecture. The structure in light yellow is \textit{Temporal module}. Best viewed in color.}
	\label{fig:dodt}
	\vspace{-0.4cm}
\end{figure*}

This paper proposes a \textbf{D}ual-way \textbf{O}bject \textbf{D}etection and \textbf{T}racking (\textbf{DODT}) framework, as illustrated in \figurename \ref{fig:dodt}, to tackle aforementioned problems. DODT is constructed based on the following observations: \textbf{(1)} Point clouds can be fused with images to obtain richer appearance features, which has been verified in \cite{chen2017multi, ku2018joint}. \textbf{(2)} Apart from 3D scene flow, temporal information can also be represented through cross-correlation between adjacent frames, as demonstrated in \cite{feichtenhofer2017detect}. \textbf{(3)} Features along consective frames change gradually, thus we can process keyframes data only and then propagate predictions to non-key frames. For (1), DODT uses a fusion scheme in \cite{ku2018joint} to aggregate features from different views. For (2), a \textit{Temporal module} is introduced for temporal information encoding through cross-correlation. Different from \cite{feichtenhofer2017detect, dosovitskiy2015flownet},  our correlation operation is performed on proposal level, which is much more flexible and efficient. For (3), DODT is designed as a dual-way fashion, thus it can operate two adjacent keyframes simultaneously. Moreover, to speed up DODT, a \textit{Shared RPN} is proposed to generate 3D proposals that can be shared by two detection branches.

To generate predictions for all frames, a \textbf{Mo}tion based \textbf{I}nterpolation (\textbf{MoI}) algorithm is developed to propagate keyframe predictions to non-key frames guided by temporal information. Meanwhile, multi-object tracking can also be accomplished through \textit{tracking by detection} \cite{lenz2015followme} paradigm. In summary, our contributions are threefold:
\begin{itemize}
	\item We propose a dual-way framework named DODT,  which performs 3D object detection and tracking precisely on streaming data based on keyframes.
	\item We propose \textit{Temporal module}, a flexible and efficient component to encode temporal information across frames in proposal level.
	\item We develop a motion based interpolation algorithm for prediction propagation, which obtain significant performance improvements on both detection and tracking.  
\end{itemize}

\section{RELATED WORK}

\textbf{3D object detection.} Currently, most approaches in 3D object detection are done in three fronts: image based, point cloud based, and multi-view fusion based. Image based approaches such as Mono3D \cite{7780605} and 3DOP \cite{chen20183d} use camera data only. Since images lack depth information, hand-crafted geometric features are required in these approaches. For point cloud based methods, there are usually two types: voxelization based and projection based, according to how point cloud features are represented. Voxelization based methods such as 3D FCN \cite{li20173d}, Vote3Deep \cite{engelcke2017vote3deep}, VoxelNet \cite{zhou2018voxelnet}, utilize a voxel grid to encode features. These approaches suffer from the sparsity of point clouds and enormous computation costs of 3D convolution. Projection based methods such as PIXOR \cite{yang2018pixor}, Complex-YOLO \cite{simon2018complex, Simon_2019_CVPR_Workshops}, attempt to project point clouds to a perspective view (e.g. bird eye view) and apply image-based feature extraction techniques. However, due to the sparsity of point clouds, features after projection are usually insufficient for accurate object detection, especially for small targets. Multi-view fusion based approaches such as F-PointNet \cite{qi2018frustum}, MV3D \cite{chen2017multi}, AVOD \cite{ku2018joint}, try to fuse point cloud features with image features for accurate object detection. These methods distinguish from each other mainly on how data is fused. Our detection branches are constructed based on AVOD, but RPN is replaced by \textit{Shared RPN}.

\textbf{Video object detection.} Nearly all existing methods in video object detection incorporate temporal information on either feature level or final box level. FGFA \cite{zhu2017flow} leverages temporal coherence on feature level, it warps the feature maps in nearby frames to a reference frame for feature enhancement according to flow emotion. On the other hand, T-CNN \cite{kang2018t, kang2016object} leverages precomputed optical flows to propagate predictions to neighboring frames; Seq-NMS \cite{han2016seq} improves NMS algorithm for video by constructing sequences along nearby high-confidence bounding boxes from consecutive frames. They all exploit temporal information in final box level. There are also a few approaches trying to learn temporal features between consecutive frames without optical flow. D\&T \cite{feichtenhofer2017detect} proposes a two branches detection network for object detection and tracking simultaneously in video. The network learns temporal information representation through computing convolutional cross-correlation between frames. Our approach is mainly inspired by D\&T. However, we develop this idea to 3D space and restrain cross-correlation in proposal level instead of a fixed window. Moreover, data association with two adjacent frames often suffers from drift and loss of targets in one frame, etc. D\&T ignores these problems, while our approach can handle these issues well by performing our MoI algorithm.

\textbf{3D multi-object tracking.} Existing 3D multi-object tracking approaches are mostly implemented based on tracking by detection paradigm. For example, FaF \cite{luo2018fast} proposes a network to predict 3D bounding boxes of past $n$ frames, and then aggregates these detections to produce accurate tracklets. 3D-CNN/PMBM \cite{scheidegger2018mono} trains a DNN to detect objects from a single image, and then feeds the detections to a Poisson multi-Bernoulli mixture tracking filter for 3D tracking. DSM \cite{frossard2018end} first predicts 3D bounding boxes in continuous frames and then associates detections using two small networks, which is similar to our approach. However, their 3D detector is single frame based approach MV3D \cite{chen2017multi}, thus temporal features between frames are mostly ignored. Moreover, their boxes association is finished by solving a linear program using a network, while ours is done by simply applying a IoU based association algorithm \cite{bochinski2018extending}.

\section{METHODOLOGY}

In this section, we first give an overview of the whole network DODT (Sec. A). Then we introduce the \textit{Shared RPN} (Sec. B) that generates shared 3D proposals. Sec. C describes how \textit{Temporal module} encodes features between two adjacent keyframes. Sec. D shows how we implement our MoI algorithm and accomplish streaming level detection as well as multi-object tracking simultaneously.

\begin{figure}
	\vspace{0.05cm}
	\begin{center}
		\includegraphics[trim={7cm, 3.5cm, 8cm, 2cm}, clip,width=0.45\textwidth]{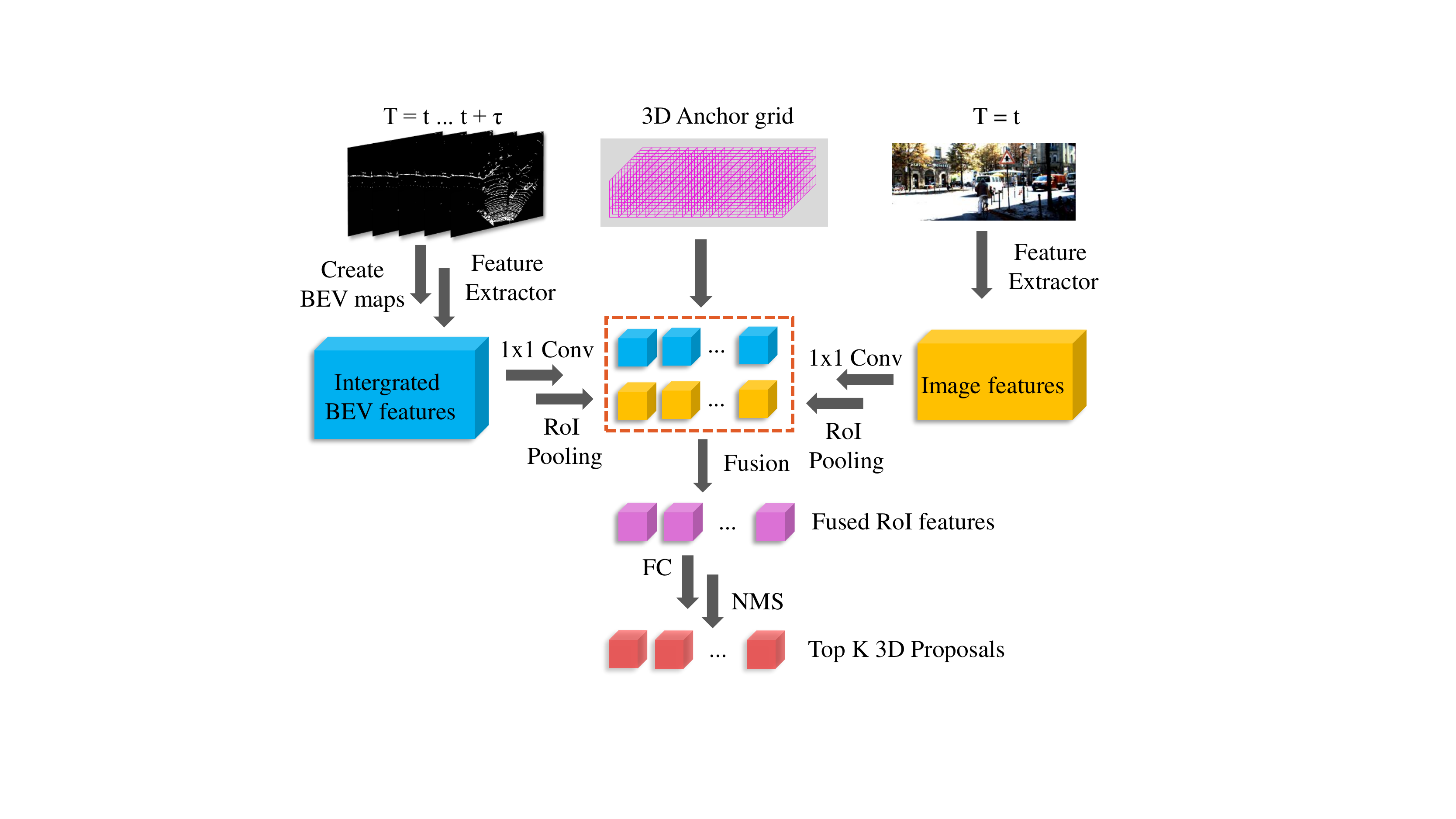}
	\end{center}
	\caption{\textit{Shared RPN} module.}
	\label{fig:rpn}
	\vspace{-0.6cm}
\end{figure}

\subsection{DODT Model Structure} 
The whole structure of DODT is illustrated in \figurename \ref{fig:dodt}. With the help of the dual-way structure, the network can be fed with two adjacent keyframes data simultaneously. Data of each frame is consist of an image and point cloud BEV maps (following the procedure described in MV3D \cite{chen2017multi}). The data is first fed to feature extractors to get corresponding feature maps. Our feature extractors are designed following the procedure described in AVOD \cite{ku2018joint}, which are constructed in an encoder-decoder fashion resulting in a full resolution feature map. To extract feature crops from image and BEV feature maps, we again follow the idea proposed in AVOD \cite{ku2018joint}. Given a 3D anchor generated by RPN, two view specific ROIs are obtained by projecting the anchor onto the image and BEV feature maps, then the corresponding feature crops are extracted and then RoI pooling is performed individually. After that, a \textit{early fusion} scheme from MV3D \cite{chen2017multi} is used to combine multi-view features in proposal level. Finally, by applying two \textit{fc} networks for classification and box regression and following 2D NMS, the final predictions are produced. Meanwhile, fed with BEV feature crops of two detection branches, \textit{Temporal module} performs correlation operation on feature crop pairs to produce cross-correlation features. These cross-correlation features are then used to predict object co-occurrence and offsets (described in Sec. C) between two keyframes for further use in MoI algorithm.

The whole DODT network is designed in an end-to-end fashion. Its multi-task objective is represented as
\begin{equation}
L = w_{cls}L_{cls} + w_{reg}L_{reg} + w_{co}L_{co} + w_{corr}L_{corr}
\end{equation}
where $L_{cls}, L_{co}$ are cross-entropy losses for classification and object co-occurrence, and $L_{reg}, L_{corr}$ are smooth \textit{L1} losses for bounding boxes regression and offsets regression. $w_{cls}, w_{reg}, w_{co}, w_{corr}$ are loss weights with values 1.0, 5.0, 1.0, 1.0, respectively. During training, $L_{reg}$ and $L_{corr}$ are normalized by the number of all proposals, while $L_{cls}, L_{co}$ are normalized by the number of positive proposals.

\subsection{Shared RPN}

\begin{figure}
	\vspace{0.05cm}
	\begin{center}
		\includegraphics[trim={5.5cm, 6cm, 9cm, 6.5cm}, clip, width=0.45\textwidth]{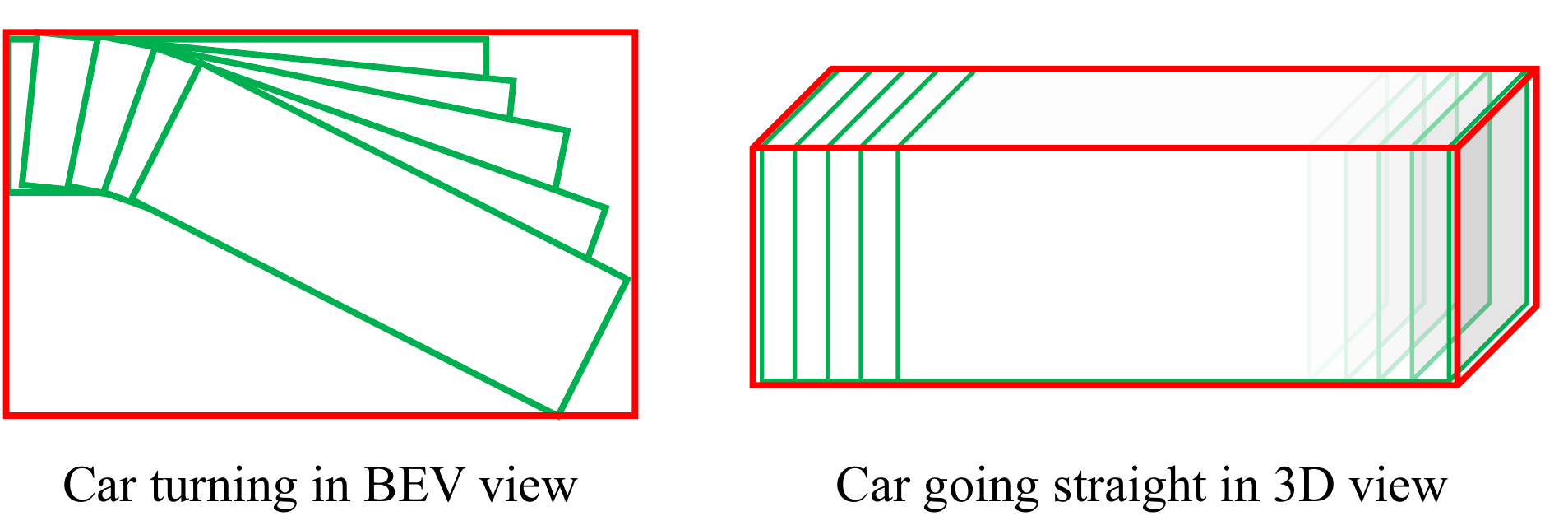}
	\end{center}
	\caption{\textcolor{green}{Green} boxes illustrate default bounding box labels of five frames, while \textcolor{red}{red} boxes are the new axis-aligned proposal labels for \textit{Shared RPN} training.}
	\label{fig:integrated_boxes}
	\vspace{-0.6cm}
\end{figure}

We transform RPN in AVOD \cite{ku2018joint} to our \textit{Shared RPN} (as shown in \figurename \ref{fig:rpn}), which can produce 3D proposals shared by both detection branches. To ensure the proposals generated by \textit{Shared RPN} are suitable for both keyframes, we create integrated BEV feature maps based on frames between two adjacent keyframes, inclusive. Note that with LIDAR poses of different timestamps, we can simply transform these frames to the same coordinate system and then integrate them. Since point cloud is extremely sparse and features are encoded by voxelization, this process will not increase any computational cost but enriches the BEV feature maps greatly. Assume that there are five frames in a frame clip, due to objects movements, the locations of the same objects in five frames are shift. For training convenience, we replace original labels of proposals with new axis-aligned labels that can cover the trajectory of an object in all five frames. \figurename \ref{fig:integrated_boxes} illustrates the new proposal labels. Though this process enlarges proposal labels slightly, it can ensure that the new labels are suitable for proposal generation in both detection branches. In addition, to obtain accurate proposals, \textit{Shared RPN} fuses BEV feature maps with image to enhance object appearance features. Since one image contains enough features in a short time, \textit{Shared RPN} only aggregates the features in the first image of a frame clip. Subsequent processes are similar with RPN module described in AVOD, we refer the reader to \cite{ku2018joint} for more details.


\subsection{Temporal Module}
Our \textit{Temporal Module} is demonstrated in the light yellow area in \figurename \ref{fig:dodt}. Given two sets of BEV features crops $F_t$ and $F_{t+\tau}$, a set of cross frame feature pairs can be constructed as $\{(F_t^i, F_{t+\tau}^i)\mid i \in \{0,1,...,N\}\}$, where $F_t^i$ and $F_{t+\tau}^i$ are features extracted by $i$-th proposal from frame $t$ and frame $t+\tau$ respectively, $\tau$ is temporal stride and $N$ is the number of 3D proposals. Note that proposals generated by \textit{Shared RPN} are shared by two detection branches, thus this set can be obtained conveniently. After the correspondence of feature crops is constructed, correlation operation can be performed on each feature pair $(F_t^i, F_{t+\tau}^i)$ to compute cross-correlation features over frames. These features are then fed to a classification head for object co-occurrence prediction and a regression head for box offsets prediction respectively. Object co-occurrence shows the probability of an object exists on both keyframes, it provides an extra guarantee for the successful implementation of our MoI algorithm. When an object is lost in one of the two keyframes, object co-occurrence determines whether a track birth or death occurs, or just a mis-detection in one keyframes. The box offsets prediction is only performed in BEV view, since vehicles usually run in a 2D plane.

For \textit{Temporal Module} training, the ground truth of a object is $b^t = (p_{co}, b^t_x, b^t_z, b^t_{ry})$ in frame $t$, and $b^{t+\tau}$ for frame $t+\tau$ similarly, denoting object co-occurrence probability, its horizontal and vertical center coordinates as well as orientation in BEV view. $p_{co}$ is 1.0 if the object exists on both keyframes, 0 otherwise. The box offsets label for regression head is $\delta^{t, t+\tau} = (\delta^{t,t+\tau}_{x}, \delta^{t,t+\tau}_{z}, \delta^{t,t+\tau}_{ry})$, represented as
\begin{equation}
\delta^{t, t+\tau} = \\
\begin{cases}
(\frac{b_{x}^{t+\tau} - b_{x}^{t}}{b^t_{w}}, \frac{b_{z}^{t+\tau} - b_{z}^{t} }{b^t_{l}}, \frac{b_{ry}^{t+\tau} - b_{ry}^{t}}{b^t_{ry}}) & p_{co} = 1.0 \\
(0.0,0.0, 0.0) &  otherwise
\end{cases}
\end{equation}
where $b^t_{w}$ and $b^t_{l}$ are width and length of the object.

\subsection{3D Streaming Object Detection and Tracking} 
MoI algorithm (as represented in Algo. \ref{alg:interpolation}) shows how we perform streaming level detection and tracking with DODT outputs. Fed with two detections lists $\{D^t, D^{t+\tau}\}$ and a offsets list $\Delta^{t, t+\tau}$ of adjacent keyframes, the algorithm first utilizes linear interpolation to calculate detections for intermediate frames if an object exists on both keyframes. Detections association between two keyframes is completed using \textit{getMatched} function.  Given a rectified detection $d_i^t + \Delta d^{t, t+ \tau}_{i}$ ($\Delta d^{t, t+ \tau}_{i}$ is decoded from $\delta^{t, t+\tau}_i$) of the first keyframe, \textit{getMatched} function select the best matched detection in $D_{temp}$ based on the 3D IoU between rectified detection and the detections in $D_{temp}$.  As for the detection that exists only in one keyframe, the match fails and the function outputs \textit{None}. In this case, if the co-occurrence probability $p_{co}^i$ is not less than $p_{co}^{max}$ (0.5 in our experiment), we regard it as a mis-detection in another keyframe, then we utilize box offsets prediction to compute corresponding detection in another keyframe and do interpolation subsequently. Otherwise we regard it as a track birth or death, then a motion model is used to propagate the detections.

The motion model holds the hypothesis that the velocity of an object is constant within a short time period, independent of camera ego-motion. Thus we can calculate object's state in next frame using the states of historical frames and the increments. In detail, we maintain a global velocity $v = (v_x, v_z, v_{ry})$ in BEV space, which will be updated to $\alpha v + (1-\alpha) \frac{\delta^{t, t+\tau}}{\tau}(b^t_w, b^t_l, b^t_{ry})$ ($\alpha = 0.8$ in our experiment) once a new detection is matched with the track. As for the end of a track, since there is no way for us to locate its exactly end frame with only keyframes predictions, we extend the track by 3 frames according to the trend of movement based on the global velocity. If a track is beyond the range of BEV feature maps, the extension will be terminated earlier. Similarly when it turns to the start of a track.

\begin{algorithm}
	\vspace{0.05cm}
	\small
	\caption{Motion based Interpolation Algorithm}
	\label{alg:interpolation}
	\textbf{Input: } $D^t= [d^t_0, d^t_1, ..., d^t_{N_t}], D^{t+\tau}= [d^{t+\tau}_0, d^{t+\tau}_1, ..., d^{t+\tau}_{N_{t+\tau}}],$
	$\Delta^{t, t+\tau}=[\delta^{t, t+\tau}_0, \delta^{t, t+\tau}_1, ..., \delta^{t, t+\tau}_{max\{N_t, N_{t+\tau}\}}]$\\
	\textbf{Output: } $D = [D^t, D^{t+1}, ..., D^{t+\tau}]$\\
	\textbf{Initialize:} $D_{temp} = D^{t+\tau}, D, p_{co}^{max} = 0.5$ \\
	\For{$d^t_i \emph{ in } D^t$}{
		$\Delta d^{t, t+ \tau}_{i} = (d^t_{i, w} \cdot \delta^{t, t+\tau}_{i,x}, 0, d^t_{i, l} \cdot \delta^{t, t+\tau}_{i,z}, 0, 0, d^t_{i, ry} \cdot \delta^{t, t+\tau}_{i,ry})$
		$d' = getMatched(d^t_i+\Delta d^{t, t+ \tau}_{i}, D_{temp})$\\
		\If{$d'$}{
			$d^{t+1}_i,..., d^{t+\tau-1}_i = Interpolate(d^t_i, d')$\\
			remove $d'$ from $D_{temp}$
		}
		\ElseIf{$p_{co}^i \geq p_{co}^{max}$}{
			$d^{t+1}_i,..., d^{t+\tau}_i = Interpolate(d^t_i, d^t_i + \Delta d^{t, t+ \tau}_{i})$
		}
		\Else{generate $(d^{t+1}_i,..., d^{t+\tau-1}_i)$ by motion model}
	}
	\If{$D_{temp}$ is not empty}{
		\For{$d^{t+\tau}_j \emph{ in } D_{temp}$}{
			\If{$p_{co}^j \geq p_{co}^{max}$}{
				$d^{t}_j,..., d^{t+\tau-1}_j = Interpolate(d^{t+\tau}_j - \Delta d^{t, t+ \tau}_{j}, d^{t+\tau}_j)$
			}
			\Else{
				generate $(d^{t+1}_j,..., d^{t+\tau-1}_j)$ by motion model
			}
		}
	}
\end{algorithm}
\setlength{\textfloatsep}{1pt}

During experiment, we observe that a track sometimes suffers from opposite orientation in nearby keyframes, which will lead to inferior performance. To address this issue, we operate a simple orientation correction algorithm during box interpolation and track extension. Consider a track $T^i_{match}$ and its next matched detection $D^i_{match}$, if the difference of orientation in $T^i_{match}$ and $D^i_{match}$ is grater than $\frac{\pi}{2}$, we add a $\pi$ to the orientation in $D^i_{match}$ so that its orientation can be roughly consistent with the track. 

Multi-object tracking can be accomplished with streaming level object detection simultaneously. Note that after performing MoI algorithm on keyframe pairs, detections association between them is also obtained. We only need to associate detections between different keyframes pairs to link tracklets over time and build long-term object tubes. Since our method associates a frame clip to a track at one time, this shows an near online tracking approach.

\section{EXPERIMENTS}
\begin{table*}\centering
	\small
	\vspace{0.15cm}
	\resizebox{\textwidth}{!}{
		\begin{tabular}{ccccccccc}
			&\multicolumn{1}{c|}{}   & \multicolumn{3}{c|}{IoU = 0.5}  		         & \multicolumn{3}{c|}{IoU = 0.7}          &  \\ \midrule
			\multicolumn{1}{c|}{Methods} & \multicolumn{1}{c|}{Modules}    & Easy     & Moderate   & \multicolumn{1}{c|}{Hard}     & Easy  & Moderate & \multicolumn{1}{c|}{Hard}    & FPS \\\midrule
			\multicolumn{1}{c|}{AVOD\cite{ku2018joint}}     &\multicolumn{1}{c|}{-}     & 90.13 / 90.91  & 80.00 / 81.79 & \multicolumn{1}{c|}{71.61 / 81.79}  & 76.00 / 90.90 & 57.23 / 81.73 & \multicolumn{1}{c|}{56.13 / 72.69}   & 10.0\\
			\multicolumn{1}{c|}{DODT($\tau$ = 1)}     &\multicolumn{1}{c|}{-}     & 88.28 / 99.97  & 85.74 / 90.90 & \multicolumn{1}{c|}{86.14 / 90.89}  & 83.44 / 90.82 & 67.48 / 90.79 & \multicolumn{1}{c|}{61.24 / 90.80}     & 6.7 \\
			\multicolumn{1}{c|}{DODT($\tau$ = 1)}     &\multicolumn{1}{c|}{T}     & 88.32 / \textbf{99.99}  & 86.53 / 90.90 & \multicolumn{1}{c|}{86.71 / \textbf{90.90}}  & 83.60 / 90.82 & 68.93 / 90.80 & \multicolumn{1}{c|}{62.69 / 90.81}   & 5.9\\
			\multicolumn{1}{c|}{DODT($\tau$ = 1)}     &\multicolumn{1}{c|}{M}     & 89.99 / 99.95  & 87.86 / 90.87 & \multicolumn{1}{c|}{87.81 / 90.86}  & 86.89 / 90.89 & 73.96 / 90.83 & \multicolumn{1}{c|}{67.07 / 81.79}   & 6.5\\
			\multicolumn{1}{c|}{DODT($\tau$ = 1)}     &\multicolumn{1}{c|}{T + M} & \textbf{90.63} / 99.95  & 89.07 / 90.90 & \multicolumn{1}{c|}{88.79 / \textbf{90.90}}  & 88.74 / 90.91 & 75.27 / 90.84 & \multicolumn{1}{c|}{68.75 / 90.57}   & 5.7\\ \midrule
			\multicolumn{1}{c|}{DODT($\tau$ = 2)}     &\multicolumn{1}{c|}{T + M} & 90.60 / 99.94  & \textbf{89.19 / 90.91} & \multicolumn{1}{c|}{\textbf{88.91} / 90.88}  & \textbf{88.90 / 90.92} & \textbf{76.64} / 90.85 & \multicolumn{1}{c|}{75.81 / 90.83}   & 8.6\\
			\multicolumn{1}{c|}{DODT($\tau$ = 3)}     &\multicolumn{1}{c|}{T + M} & 90.61 / 99.98  & 89.01 / 90.89 & \multicolumn{1}{c|}{88.84 / 90.89}  & 88.81 / 90.91 & 76.38 / \textbf{90.86} & \multicolumn{1}{c|}{\textbf{75.83 / 90.85}}   & 11.4\\
			\multicolumn{1}{c|}{DODT($\tau$ = 4)}     &\multicolumn{1}{c|}{T + M} & 90.55 / 99.94  & 88.82 / 90.88 & \multicolumn{1}{c|}{88.34 / 90.87}  & 88.43 / 90.91 & 75.70 / 90.82 & \multicolumn{1}{c|}{68.75 / 90.82}   & 14.3\\
			\multicolumn{1}{c|}{DODT($\tau$ = 5)}     &\multicolumn{1}{c|}{T + M} & 87.98 / 90.91  & 85.57 / 90.87 & \multicolumn{1}{c|}{86.01 / 90.87}  & 81.59 / 90.81 & 67.30 / 90.76 & \multicolumn{1}{c|}{61.35 / 81.73}   & 17.1\\
			\multicolumn{1}{c|}{DODT($\tau$ = 6)}     &\multicolumn{1}{c|}{T + M} & 78.77 / 90.75  & 70.88 / 90.71 & \multicolumn{1}{c|}{71.65 / 81.70}  & 71.71 / 90.44 & 55.86 / 81.50 & \multicolumn{1}{c|}{56.80 / 81.51}   & \textbf{20.0} \\ \midrule
	\end{tabular}}
	\setlength{\abovecaptionskip}{-1pt}
	\caption*{TABLE II: We report $AP_{3D}/AP_{BEV}$ (in \%) of the \textbf{Car} category on validation datasets, corresponding to average precision of 3D object detection in 3D view and in BEV view. T is \textit{Temporal Module}, M is our MoI algorithm. $\tau$ is temporal stride.} 
	\setlength{\belowcaptionskip}{1pt}
	\label{table:result_detection}
	\vspace{-0.65cm}
\end{table*}

\subsection{Datasets and Training}

\textbf{Datasets and Preprocessing.} We use KITTI object tracking Benchmark \cite{geiger2013vision} for evaluation. It consists of 21 training sequences and 29 test sequences with vehicles annotated in 3D. For object detection, we split 21 training sequences into two parts according to their sequence number, odd numbered sequences for training and the rest for validation. For tracking evaluation, we train our model in all 21 training sequences and evaluate on test datasets. Similar to the data preprocessing in AVOD \cite{ku2018joint}, we crop point clouds at $[-40, 40] \times [0, 70] \times [0, 2.5]$ meters along $X, Z, Y$ axis respectively to contain points within the field of camera view. In addition, to make the system invariant to the speed of the ego-car, we calculate the displacement of the observer between adjacent keyframes using IMU data provided by KITTI, and then transform the coordinates in second keyframe accordingly. 

We notice that the labels provided in KITTI object tracking datasets are incomplete. For example, first row in \figurename \ref{fig:examples} illustrates several frames and labels in training sequence 0. We can see that some objects (marked by red rectangles) are not labeled, even though they can be well observed in frame 118 and 120 and are labeled in frame 128. However, our model can well predict these objects, as shown in second row. In order to evaluate our model as accurately as possible, we add these missing labels manually.

\textbf{Training and testing.} We train DODT only for \textit{Car} category temporarily, following most of the super-parameter settings in AVOD \cite{ku2018joint} during training and testing. The network is trained for 120K iterations with a batch size 1, using an ADAM \cite{kingma2014adam} optimizer with an initial learning rate of 0.0001 that is decayed exponentially every 30K iterations with a decay factor of 0.8. During proposal generation, anchors with IoU less than 0.3 are considered background and greater than 0.5 are objects. To remove redundant proposals, 2D NMS is performed at an IoU threshold of 0.8 in BEV to keep the top 1024 proposals during training, while at inference time, the top 300 proposals are kept. 

\subsection{Results}
\textbf{Shared RPN.} To evaluate the performance of \textit{Shared RPN}, we implement a non-shared version of RPN. It predicts proposals for each keyframe independently based on each feature maps. A comparison of proposal prediction accuracy between \textit{Non-shared RPN} and \textit{Shared RPN} is shown in \tablename \, I. Results show that \textit{Shared RPN} outperforms \textit{Non-shared RPN} by 0.66\%, which indicates that the shared mechanism in RPN promotes the accuracy of proposal prediction. 
\begin{table}[h]\centering
	\vspace{-0.5cm}
		\begin{tabular}{ccc}
			\toprule[1pt]
			Method        & Non-shared RPN & Shared RPN  \\ \midrule
			Accuracy(\%)  & 97.81      & \textbf{98.47}       \\
			\bottomrule[1pt]
	\end{tabular}
	\caption*{TABLE I: Comparison of proposal prediction accuracy.}
	\label{table:rpn_result}
	\vspace{-0.4cm}
\end{table}

\textbf{Streaming level detection.} The main results of 3D object detection are summarized in \tablename \, II. We first evaluate the effectiveness of the dual-way structure, \textit{Temporal Module} and our MoI algorithm in 3D object detection with a temporal stride $\tau = 1$. Several important trends can be observed: \textbf{1)} Compared with original AVOD \cite{ku2018joint} model trained on the same datasets, our base model (without \textit{Temporal Module} and MoI algorithm) shows improvements in all settings with IoU = 0.7. These improvements indicate that the dual-way structure with \textit{Shared RPN} contributes to detection performance, as it can aggregate features in consecutive frames to produce more accurate proposals. \textbf{2)} The introduction of \textit{Temporal Module} and MoI algorithm is conducive to model performance. Compared with our base model, \textit{Temporal Module} brings 0.16\%, 1.45\% and 1.45\% gain in \textit{Easy}, \textit{Moderate} and \textit{Hard} setting respectively with IoU = 0.7. It shows that the aid of object co-occurrences information contributes to the detection in occluded and truncated objects more than normal ones. The comparison also shows that our MoI algorithm improves model accuracy by 3.45\%, 6.48\% and 5.83\% in three setting respectively with a overlap of 0.7. These improvements indicate that MoI algorithm works well in all settings. It's not surprising because the algorithm has the ability to reject false positive predictions and generate new results by extending the trajectories. \textbf{3)} \textit{Temporal Module} and MoI algorithm can work synergistically and improve model accuracy by 1-2\% additionally in all setting, since our MoI algorithm can work better with the outputs of \textit{Temporal Module}. 

\begin{figure*}\centering
	\vspace{0.15cm}
	\begin{center}
		\includegraphics[trim={2cm, 1cm, 2.5cm, 1cm}, clip, width=\textwidth]{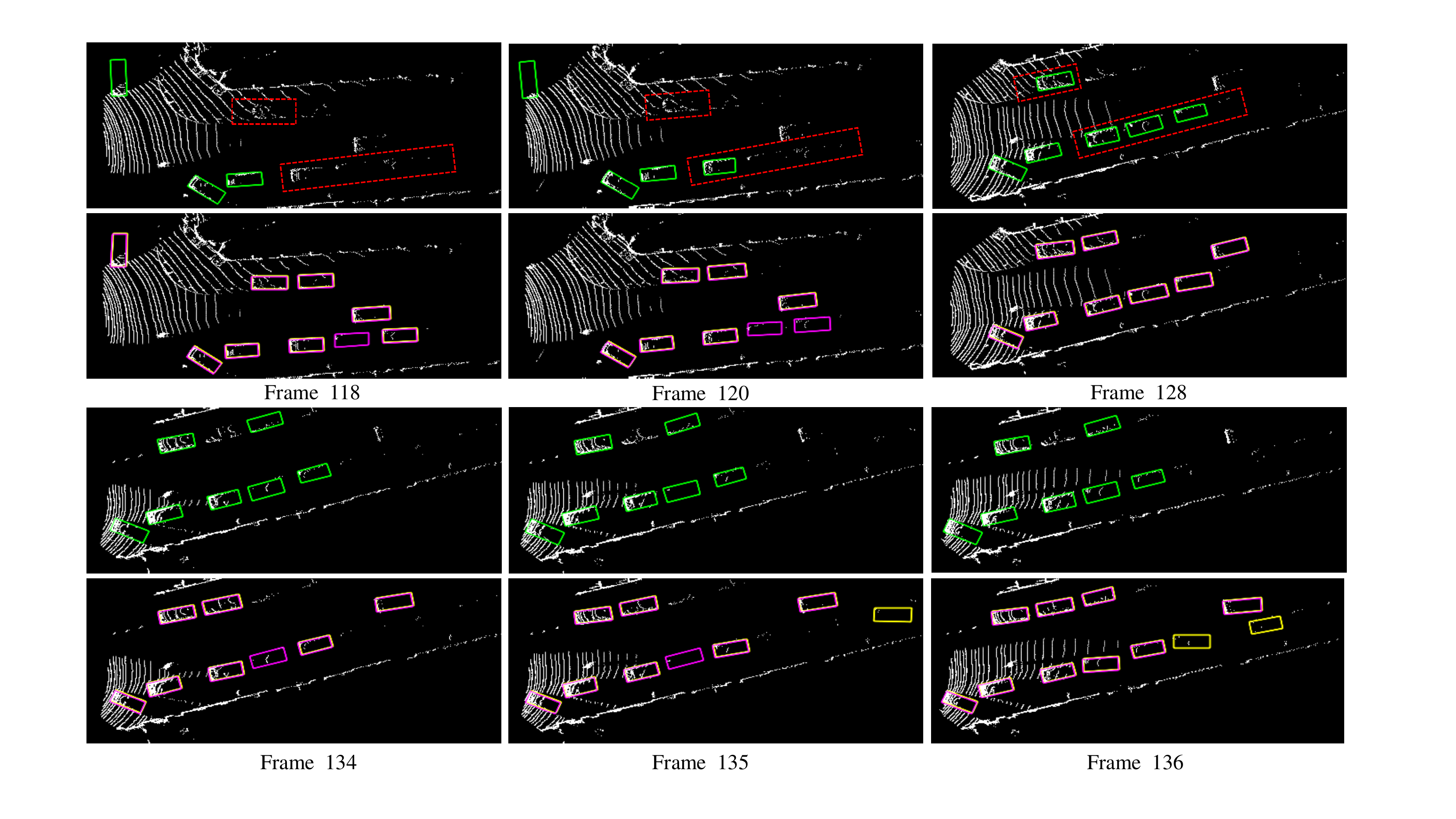}
	\end{center}
	\setlength{\abovecaptionskip}{-3pt} 
	\caption{Visualization of labels and our predictions in sequences 0. Boxes in \textcolor{green}{green} are official ground truth, boxes in \textcolor{yellow}{yellow} are results predicted with temporal stride $\tau = 1$, and boxes in \textcolor{magenta}{magenta} are results predicted with temporal stride $\tau = 3$. Boxes in mixed color are magenta boxes overlapped with yellow boxes. Best viewed in color.}
	\label{fig:examples}
	\vspace{-0.6cm}
\end{figure*}

Secondly, we investigate the effect of temporal stride $\tau$ on model performance. We train six models with $\tau = \{1, 2, 3, 4, 5, 6\}$ respectively and then perform MoI algorithm to generate predictions for all frames. Results are shown in \tablename \, II. DODT ($\tau = 2$) shows the best result in \textit{Easy} and \textit{Moderate} settings with IoU = 0.7, by 88.90\% and 76.64\% in $AP_{3D}$ respectively. DODT ($\tau = 3$) performs the best in \textit{Hard} setting, by 75.84\% in $AP_{3D}$. Compared with DODT ($\tau = 1$), model performance with $\tau = 2, 3$ are boosted in hard objects, by about 1\% in \textit{Moderate} setting and about 7\% in \textit{Hard} setting with IoU = 0.7. These gains demonstrate that the detection of truncated and occluded objects can benefit greatly from a large temporal stride. Bottom two rows in \figurename \ref{fig:examples} shows how MoI algorithm improves model performance. Boxes in magenta are DODT predictions with $\tau = 3$ while yellow with $\tau = 1$. these results shows that a larger $\tau$ can reduce noisy detections and generate more true positive predictions. However,  \tablename \, II also shows that a too large $\tau$ leads to a significant decay of accuracy. As temporal stride increasing, there are more tracks start or end within two adjacent keyframes, thus our MoI algorithm would fail to generate most objects near the start or end of a track. Moreover, temporal information in a longer time period is more difficult to represent.

We also compute the inference time in streaming level detection. The total runtime of two detection branches is 175ms on a Tesla P100 GPU. As temporal stride $\tau$ increasing, the time cost for detection is unchanged, and interpolation time cost increases negligibly, thus the total inference time is almost unchanged, but FPS increases by multiple. We choose $\tau = 3$ for our following experiment, which is a good trade-off between speed and accuracy.

\textbf{Multi-object tracking.} For multi-object tracking, We investigate the contribution of different modules to tracking accuracy in KITTI tracking validation datasets. Results in \tablename \, III shows that \textbf{1)} our approach outperforms original AVOD model by a large margin (e.g. MOTA by 13.67\%, MOTP by 0.58\%, MT by 25.63\%, etc.). These improvements mainly benefit from better object detection results. \textbf{2)} Compared with base model, performances of the model aided with two modules are superior in nearly all tracking metrics, as \textit{Temporal Module} makes the data association more accurate, and box interpolation helps to complete both ends of tracks. 

We also compare our approach to publicly available methods of multiple object tracking in 3D on KITTI Tracking Benchmark. Results are shown in \tablename \, IV. We can see that our approach is competitive with the state-of-the-art in some metrics (e.g. IDS and FM). For multiple object tracking accuracy, our method outperforms Complexer-YOLO \cite{Simon_2019_CVPR_Workshops} and DSM \cite{frossard2018end}, but behind 3D-CNN/PMBM \cite{scheidegger2018mono} and 3DT \cite{Hu3DT19}. For trajectory ID-switches and fragmentations, our method outperforms all other methods. Note that 3D-CNN/PMBM utilizes PMBM filter for data association while ours is a simple IoU based method, and 3DT trained their system additionally on new datasets collected on a realistic 3D virtual environments. Moreover, labels of test datasets may be incomplete just like the case in training datasets, thus the tracking performance of our approach may be better than current ones. In addition, our approach shows a competitive results in runtime with 76.9 FPS (on a  3.2 GHz Intel CPU). Note that we could not separate predictions propagation from data association easily in Algo. \ref{alg:interpolation}, thus the tracking runtime also includes boxes interpolation process.
\begin{table}
	\resizebox{0.48\textwidth}{!}{
		\begin{tabular}{cccccccc}
			\toprule[1pt]
			Methods   & Modules & MOTA(\%)$\uparrow$ & MOTP(\%)$\uparrow$ & MT(\%)$\uparrow$ & ML(\%)$\downarrow$ & IDS$\downarrow$&  FM$\downarrow$ \\ \midrule
			AVOD\cite{ku2018joint}    & -      & 66.05    & 82.97    & 46.22  & 12.18  & \textbf{2}      &  113  \\
			DODT($\tau$ = 3)          & -      & 76.53    & \textbf{83.93}    & 68.91  & 7.14   & 32      &  80  \\
			DODT($\tau$ = 3)          & T      & 77.52    & 83.75    & 69.33  & 7.56   & 37     &  77  \\
			DODT($\tau$ = 3)          & M      & 78.73    & \textbf{83.93}    & 68.49  & 9.55   & \textbf{2}      &  \textbf{48}  \\
			DODT($\tau$ = 3)  & T+M    & \textbf{79.72}   & 83.55    & \textbf{71.85}  & \textbf{5.46}  & 7  &  66  \\ 
			\bottomrule[1pt]
	\end{tabular}}
	\caption*{TABLE III: Ablation study on KITTI Tracking validation datasets.}
	\label{label:result_tracking}
\end{table}
\begin{table}
	\vspace{-0.2cm}
	\resizebox{0.48\textwidth}{!}{
		\begin{tabular}{cccccccc}
			\toprule[1pt]
			Method    & MOTA(\%)$\uparrow$ & MOTP(\%)$\uparrow$ & MT(\%)$\uparrow$ & ML(\%)$\downarrow$ & IDS$\downarrow$&  FM$\downarrow$ &FPS$\uparrow$ \\ \midrule
			Complexer-YOLO\cite{Simon_2019_CVPR_Workshops}    & 75.70    & 78.46    & 58.00  & 5.08  & 1186 & 2096 & \textbf{100.0} \\ 
			DSM\cite{frossard2018end}                         & 76.15    & 83.42    & 60.00  & 8.31  & 296  & 868  & 10.0 (GPU)  \\ 
			3D-CNN/PMBM\cite{scheidegger2018mono}             & 80.39    & 81.26	& 62.77  & 6.15  & 121  & 613  & 71.4 \\  
			3DT\cite{Hu3DT19} 	                              & \textbf{84.52}    & \textbf{85.64}	& \textbf{73.38}  & \textbf{2.77}  & 377  & 847  & 33.3 \\ 
			DODT(ours)                                        & 76.68    & 81.65    & 60.77  & 11.69 & \textbf{63}   & \textbf{384}  & 76.9 \\ 
			\bottomrule[1pt]
	\end{tabular}}
	\caption*{TABLE IV: Comparison of publicly available methods of 3D multi-object tracking in the KITTI Tracking Benchmark. The time for object detection is not included in the specified runtime.}
	\label{label:result_kitti}
\end{table}
\section{CONCLUSIONS}
\label{sec:conclusions} We propose DODT, a unified framework for simultaneous 3D object detection and tracking based on streaming data. The network is a dual-way structure and can  process two keyframes at the same time. Embedded with a \textit{Temporal Module} to encode the temporal information among adjacent keyframes and a motion based interpolation algorithm to propagate predictions to non-key frames, our network can perform object detection and tracking in a very efficient way.
\section{ACKNOWLEDGEMENTS}
This work has been partly funded by the Shenzhen Science and Technology Innovation Committee under Grant NO.JCYJ20180507182508857, and the Ministry of Science and Technology of the People's Republic of China under Grant NO.2018YFB1802400.
\bibliographystyle{IEEEtran}
\bibliography{egbib}

\begin{thebibliography}{10}
\providecommand{\url}[1]{#1}
\csname url@samestyle\endcsname
\providecommand{\newblock}{\relax}
\providecommand{\bibinfo}[2]{#2}
\providecommand{\BIBentrySTDinterwordspacing}{\spaceskip=0pt\relax}
\providecommand{\BIBentryALTinterwordstretchfactor}{4}
\providecommand{\BIBentryALTinterwordspacing}{\spaceskip=\fontdimen2\font plus
\BIBentryALTinterwordstretchfactor\fontdimen3\font minus
  \fontdimen4\font\relax}
\providecommand{\BIBforeignlanguage}[2]{{%
\expandafter\ifx\csname l@#1\endcsname\relax
\typeout{** WARNING: IEEEtran.bst: No hyphenation pattern has been}%
\typeout{** loaded for the language `#1'. Using the pattern for}%
\typeout{** the default language instead.}%
\else
\language=\csname l@#1\endcsname
\fi
#2}}
\providecommand{\BIBdecl}{\relax}
\BIBdecl

\bibitem{7780605}
X.~{Chen}, K.~{Kundu}, Z.~{Zhang}, H.~{Ma}, S.~{Fidler}, and R.~{Urtasun},
  ``Monocular 3d object detection for autonomous driving,'' in \emph{2016 IEEE
  Conference on Computer Vision and Pattern Recognition (CVPR)}, June 2016, pp.
  2147--2156.

\bibitem{chen20183d}
X.~Chen, K.~Kundu, Y.~Zhu, H.~Ma, S.~Fidler, and R.~Urtasun, ``3d object
  proposals using stereo imagery for accurate object class detection,''
  \emph{IEEE transactions on pattern analysis and machine intelligence},
  vol.~40, no.~5, pp. 1259--1272, 2018.

\bibitem{zhou2018voxelnet}
Y.~Zhou and O.~Tuzel, ``Voxelnet: End-to-end learning for point cloud based 3d
  object detection,'' in \emph{Proceedings of the IEEE Conference on Computer
  Vision and Pattern Recognition}, 2018, pp. 4490--4499.

\bibitem{yang2018pixor}
B.~Yang, W.~Luo, and R.~Urtasun, ``Pixor: Real-time 3d object detection from
  point clouds,'' in \emph{Proceedings of the IEEE Conference on Computer
  Vision and Pattern Recognition}, 2018, pp. 7652--7660.

\bibitem{simon2018complex}
M.~Simon, S.~Milz, K.~Amende, and H.-M. Gross, ``Complex-yolo: An
  euler-region-proposal for real-time 3d object detection on point clouds,'' in
  \emph{European Conference on Computer Vision}.\hskip 1em plus 0.5em minus
  0.4em\relax Springer, 2018, pp. 197--209.

\bibitem{chen2017multi}
X.~Chen, H.~Ma, J.~Wan, B.~Li, and T.~Xia, ``Multi-view 3d object detection
  network for autonomous driving,'' in \emph{Proceedings of the IEEE Conference
  on Computer Vision and Pattern Recognition}, 2017, pp. 1907--1915.

\bibitem{ku2018joint}
J.~Ku, M.~Mozifian, J.~Lee, A.~Harakeh, and S.~L. Waslander, ``Joint 3d
  proposal generation and object detection from view aggregation,'' in
  \emph{2018 IEEE/RSJ International Conference on Intelligent Robots and
  Systems (IROS)}.\hskip 1em plus 0.5em minus 0.4em\relax IEEE, 2018, pp. 1--8.

\bibitem{feichtenhofer2017detect}
C.~Feichtenhofer, A.~Pinz, and A.~Zisserman, ``Detect to track and track to
  detect,'' in \emph{Proceedings of the IEEE International Conference on
  Computer Vision}, 2017, pp. 3038--3046.

\bibitem{dosovitskiy2015flownet}
A.~Dosovitskiy, P.~Fischer, E.~Ilg, P.~Hausser, C.~Hazirbas, V.~Golkov, P.~Van
  Der~Smagt, D.~Cremers, and T.~Brox, ``Flownet: Learning optical flow with
  convolutional networks,'' in \emph{Proceedings of the IEEE international
  conference on computer vision}, 2015, pp. 2758--2766.

\bibitem{lenz2015followme}
P.~Lenz, A.~Geiger, and R.~Urtasun, ``Followme: Efficient online min-cost flow
  tracking with bounded memory and computation,'' in \emph{Proceedings of the
  IEEE International Conference on Computer Vision}, 2015, pp. 4364--4372.

\bibitem{li20173d}
B.~Li, ``3d fully convolutional network for vehicle detection in point cloud,''
  in \emph{2017 IEEE/RSJ International Conference on Intelligent Robots and
  Systems (IROS)}.\hskip 1em plus 0.5em minus 0.4em\relax IEEE, 2017, pp.
  1513--1518.

\bibitem{engelcke2017vote3deep}
M.~Engelcke, D.~Rao, D.~Z. Wang, C.~H. Tong, and I.~Posner, ``Vote3deep: Fast
  object detection in 3d point clouds using efficient convolutional neural
  networks,'' in \emph{2017 IEEE International Conference on Robotics and
  Automation (ICRA)}.\hskip 1em plus 0.5em minus 0.4em\relax IEEE, 2017, pp.
  1355--1361.

\bibitem{Simon_2019_CVPR_Workshops}
M.~Simon, K.~Amende, A.~Kraus, J.~Honer, T.~Samann, H.~Kaulbersch, S.~Milz, and
  H.~Michael~Gross, ``Complexer-yolo: Real-time 3d object detection and
  tracking on semantic point clouds,'' in \emph{The IEEE Conference on Computer
  Vision and Pattern Recognition (CVPR) Workshops}, June 2019.

\bibitem{qi2018frustum}
C.~R. Qi, W.~Liu, C.~Wu, H.~Su, and L.~J. Guibas, ``Frustum pointnets for 3d
  object detection from rgb-d data,'' in \emph{Proceedings of the IEEE
  Conference on Computer Vision and Pattern Recognition}, 2018, pp. 918--927.

\bibitem{zhu2017flow}
X.~Zhu, Y.~Wang, J.~Dai, L.~Yuan, and Y.~Wei, ``Flow-guided feature aggregation
  for video object detection,'' in \emph{Proceedings of the IEEE International
  Conference on Computer Vision}, 2017, pp. 408--417.

\bibitem{kang2018t}
K.~Kang, H.~Li, J.~Yan, X.~Zeng, B.~Yang, T.~Xiao, C.~Zhang, Z.~Wang, R.~Wang,
  X.~Wang \emph{et~al.}, ``T-cnn: Tubelets with convolutional neural networks
  for object detection from videos,'' \emph{IEEE Transactions on Circuits and
  Systems for Video Technology}, vol.~28, no.~10, pp. 2896--2907, 2018.

\bibitem{kang2016object}
K.~Kang, W.~Ouyang, H.~Li, and X.~Wang, ``Object detection from video tubelets
  with convolutional neural networks,'' in \emph{Proceedings of the IEEE
  conference on computer vision and pattern recognition}, 2016, pp. 817--825.

\bibitem{han2016seq}
W.~Han, P.~Khorrami, T.~L. Paine, P.~Ramachandran, M.~Babaeizadeh, H.~Shi,
  J.~Li, S.~Yan, and T.~S. Huang, ``Seq-nms for video object detection,''
  \emph{arXiv preprint arXiv:1602.08465}, 2016.

\bibitem{luo2018fast}
W.~Luo, B.~Yang, and R.~Urtasun, ``Fast and furious: Real time end-to-end 3d
  detection, tracking and motion forecasting with a single convolutional net,''
  in \emph{Proceedings of the IEEE Conference on Computer Vision and Pattern
  Recognition}, 2018, pp. 3569--3577.

\bibitem{scheidegger2018mono}
S.~Scheidegger, J.~Benjaminsson, E.~Rosenberg, A.~Krishnan, and
  K.~Granstr{\"o}m, ``Mono-camera 3d multi-object tracking using deep learning
  detections and pmbm filtering,'' in \emph{2018 IEEE Intelligent Vehicles
  Symposium (IV)}.\hskip 1em plus 0.5em minus 0.4em\relax IEEE, 2018, pp.
  433--440.

\bibitem{frossard2018end}
D.~Frossard and R.~Urtasun, ``End-to-end learning of multi-sensor 3d tracking
  by detection,'' in \emph{2018 IEEE International Conference on Robotics and
  Automation (ICRA)}.\hskip 1em plus 0.5em minus 0.4em\relax IEEE, 2018, pp.
  635--642.

\bibitem{bochinski2018extending}
E.~Bochinski, T.~Senst, and T.~Sikora, ``Extending iou based multi-object
  tracking by visual information,'' \emph{AVSS. IEEE}, 2018.

\bibitem{geiger2013vision}
A.~Geiger, P.~Lenz, C.~Stiller, and R.~Urtasun, ``Vision meets robotics: The
  kitti dataset,'' \emph{The International Journal of Robotics Research},
  vol.~32, no.~11, pp. 1231--1237, 2013.

\bibitem{kingma2014adam}
D.~P. Kingma and J.~Ba, ``Adam: A method for stochastic optimization,''
  \emph{arXiv preprint arXiv:1412.6980}, 2014.

\bibitem{Hu3DT19}
H.-N. Hu, Q.-Z. Cai, D.~Wang, J.~Lin, M.~Sun, P.~Krähenbühl, T.~Darrell, and
  F.~Yu, ``Joint monocular 3d detection and tracking,'' 2019.

\end{thebibliography}
\end{document}